%% file: navigation.tex
\documentclass[letterpaper, 10 pt, conference]{ieeeconf}  % Comment this line out if you need a4paper
\usepackage[]{graphicx}
\usepackage[abs]{overpic}
\usepackage{amsmath}
\usepackage{amsfonts}
\usepackage{amssymb}
\usepackage{multirow}
\usepackage{color}

\IEEEoverridecommandlockouts                              % This command is only needed if
\title{\LARGE \bf
Reinforcement Learning-based Visual Navigation with Information-Theoretic Regularization}

\author{Qiaoyun~Wu,
        Kai~Xu\thanks{Corresponding author: Kai Xu (kevin.kai.xu@gmail.com)},
        Jun~Wang\thanks{Corresponding author: Jun Wang (wjun@nuaa.edu.cn)},
        Mingliang Xu,
        Xiaoxi Gong,
        Dinesh Manocha
\thanks{Q.~Wu, J.~Wang, X.~Gong are with
Nanjing University of Aeronautics and Astronautics, China.}% <-this % stops a space
\thanks{K.~Xu is with the School of Computer Science, National University of Defense Technology, China.}
\thanks{M.~Xu is with the Zhengzhou University, China.}
\thanks{D. Manocha is with the Department of Computer Science, the University of Maryland, College Park.}
\thanks{Digital Object Identifier (DOI): 10.1109/LRA.2020.3048668}
}

\begin{document}

\maketitle
\thispagestyle{empty}
\pagestyle{empty}

%%%%%%%%%%%%%%%%%%%%%%%%%%%%%%%%%%%%%%%%%%%%%%%%%%%%%%%%%%%%%%%%%%%%%%%%%%%%%%%%
\begin{abstract}
To enhance the cross-target and cross-scene generalization of target-driven visual navigation based on deep reinforcement learning (RL),
we introduce an information-theoretic regularization term into the RL objective.
The regularization maximizes the mutual information between navigation actions
and visual observation transforms of an agent, thus promoting more informed navigation decisions.
This way, the agent models the action-observation dynamics by learning a variational generative model. Based on the model, the agent generates (imagines) the next observation from its current observation and navigation target.
This way, the agent learns to understand the causality between navigation actions and the changes in its observations,
which allows the agent to predict the next action for navigation by comparing the current and the imagined next observations.
Cross-target and cross-scene evaluations on the AI2-THOR framework  show that our method attains at least $10\%$ improvement of average success rate over some state-of-the-art models.
We further evaluate our model in two real-world settings:
navigation in unseen indoor scenes from a discrete Active Vision Dataset (AVD) and continuous real-world environments with a TurtleBot.
We demonstrate that our navigation model is able to successfully achieve navigation tasks in
these scenarios.
\emph{Videos and models can be found in the supplementary material.}\footnote{https://ieeexplore.ieee.org/document/9312496}
\end{abstract}

\input{introduction}

\input{related}
\input{method}

%%-------------------------------------------------------------------------
\input{result_discuss}

%%-------------------------------------------------------------------------
\input{conclusion}

\input{appendix}

\bibliographystyle{IEEEtran}

\bibliography{egbib}

\end{document}

%% file: introduction.tex
\section{Introduction}\vspace{-4pt}
Visual navigation is one of the basic components necessary for an autonomous agent to perform a variety of tasks in complex environments.
This component can be characterized as the ability of an agent to
understand its surrounding environments and navigate efficiently and safely to a designated target solely based on  input from on-board visual sensors~\cite{zhu2017,yang2018visual,wortsman2019,bansal2020combining}.
This encompasses two key points.
First, the agent should be able to analyze and infer the aspects most relevant to the target from the current observation to guide the decision.
Second, the agent should understand the correlation and causality between navigation actions and the changes in its observation of the surroundings.

%
%
%Most existing robotic navigation approaches tend to perform 3D metric and semantic mapping before path planning and %control~\cite{davison1998,thrun2005,bachrach2009,zhang2012,schmid2013,ccelik2013}. Such approaches are sensitive to %noisy sensory inputs and changes in the environment.
Recently, there has been an increased interest in mapless visual navigation approaches where the agent neither relies on the prior knowledge of the environment, nor performs online mapping.
Instead, it predicts navigation actions directly from observational pixels thanks to end-to-end deep learning, e.g., Imitation Learning~\cite{Pomerleau1993,ross2011,sun2017deeply,zhu2018scores}  and Deep Reinforcement Learning~\cite{yang2018visual,mirowski2016,hsu2018distributed}.
Despite significant progress in visual navigation, the generalization to novel targets and unseen scenes is still a fundamental challenge.
%~\cite{gordon2019,savva2019habitat,wu2019bayesian,zhu2019sim}
%In these tasks, CNNs are typically trained for feature extraction from observation images and fully connected layers map the features to a probability distribution over actions. Thus, these can learn navigation behaviors by leveraging extensive navigation experience in similar environments.
%However, most of approaches are based on carefully designed architectures and have only been demonstrated to perform well in simple synthetic scenes~\cite{mnih2016,oh2016control,dosovitskiy2016learning}.
In this work, we focus on visual navigation, driven by targets represented by an image, with both cross-scene and cross-target generalization.

To achieve visual navigation, we propose enhancing a Deep Reinforcement Learning approach (e.g. A3C~\cite{zhu2017}) with an information-theoretic regularization. We introduce the regularization into the RL objective to guide the agent in a more informative search for its navigation actions.
In particular, the regularization maximizes the mutual information between the action and the next visual observation given the current visual observation of the agent. This way, the agent models the action-observation dynamics and learns to understand the causality between navigation actions and the changes in its observations, thus making more informed decisions.

The maximization is, however, intractable due to the unknown next visual observation at each time step. Inspired by our previous work~\cite{wuneonav}, which presents a variational Bayesian model (NeoNav) for supervised navigation learning\footnote{https://github.com/wqynew/NeoNav}, we introduce a variational auto-encoder (VAE) model, which generates (imagines) the next observation based on the current observation and the target view.
We regularize the latent space of the VAE through the action-observation dynamics.
The agent then learns to predict the next action based on the current and the \emph{imagined} observations.
Consequently, the agent essentially builds a connection between the current observation and the target to infer the most relevant part for navigation and makes decisions based on the causality between navigation actions and observational changes.

There are several works on introducing information-theoretic regularization to RL~\cite{liu2019policy,yang2019regularized}.
Most of them strive to maximize the entropy of the policy to encourage exploration or to make the policy more stochastic for better robustness.
A specifically related work is~\cite{mohamed2015variational}, which devises a similar mutual information maximization as an internal reward for learning an intrinsically motivated agent.
In contrast to their work, we use mutual information maximization as a regularization of the objective and learn a generative model of the action-observation dynamics.
To our knowledge, our method is the first to use information-theoretic regularization to guide the learning of generalizable visual navigation.

%The approach in this work stems from our previous work~\cite{wuneonav}, which presents a variational Bayesian model (NeoNav) for supervised navigation learning\footnote{https://github.com/wqynew/NeoNav}. Here we incorporate this generative model into a deep RL framework by introducing an information-theoretic regularization to further improve the navigation performance in novel scenes.

In summary, our contributions are as follows:
(1) We present a novel method of  incorporating supervision into an RL framework (A3C) by introducing an information-theoretic regularization for improving the sample efficiency in target-driven visual navigation policy learning.
(2) We propose a visual navigation model that builds the causality between robot actions and visual transformations, and the connection between visual observations and navigation targets. This significantly improves robot navigation performance in unseen environments with novel targets.
We conduct extensive evaluations on datasets from both synthetic and real-world scenes, including AI2-THOR~\cite{zhu2017} and AVD~\cite{mousavian2019visual}.
Our model outperforms some state-of-the-art methods significantly (e.g., at least $10\%$ higher success rate for both cross-target and cross-scene evaluation on AI2-THOR). Furthermore, we show that our model, trained on the discrete household dataset (e.g., AVD) and deployed on a Turtlebot, can transfer to realistic public scenes
and exhibit robustness towards the target type and the scene layout.
%\emph{Our code has been submitted and will be made publicly available.}

%% file: related.tex
\section{Related Works}
Autonomous navigation in an unknown environment is one of the core problems in mobile  robotics and it has been extensively studied.
In this section, we provide a brief overview of some relevant works.

\textbf{Reinforcement Learning.}
Recently, a growing number of methods have been reported for RL-based navigation~\cite{zhu2017,mirowski2016,chen2017,zhang2017,kahn2018self,xia2020interactive}.
For example, Jaderberg et al.~\cite{jaderberg2016} take advantage of auxiliary control or reward prediction tasks to assist reinforcement learning in synthetic 3D maze environments.
Direct prediction of future measurements during learning also appears effective for sensorimotor control in simple immersive environments~\cite{dosovitskiy2016learning}.
Gupta et al.~\cite{gupta2017} present an end-to-end architecture to jointly train mapping and planning for navigation in novel scenes with the perfect odometry available assumption.
Savinov et al.~\cite{savinov2018semi} propose the use of topological graphs for the task of navigation and require several minutes of footage before navigating in an unseen scene.
Kahn et al.~\cite{kahn2018self} explore the intersection between model free algorithms and model-based algorithms in the context of learning navigation policies.
Wei et al.\cite{yang2018visual} integrate semantic and functional priors to improve navigation performance and can generalize to unseen scenes and objects.
Xie et al.~\cite{trigonisnapnav} propose using a few snapshots of the environment combined with directional guidance to help execute navigation tasks.
Hirose et al.~\cite{hirose2019deep} introduce a learning agent that
can follow a demonstrated path. The path consists of raw image sequences when navigating in an environment which largely discounts the practicality.

\textbf{Combined Learning Methods.}
Methods combining the advantages of imitation learning (IL) and RL have become popular ~\cite{gao2018reinforcement,gimelfarb2018reinforcement,zhu2018reinforcement}. These works provide suitable expert demonstrations to mitigate the low RL sample efficiency problem.
Ho et al.~\cite{ho2016generative} exploit a generative adversarial model to fit distributions of states and actions defining expert behavior.
They learn a policy from supplied data and hence avoid the costly expense of RL.
~\cite{li2018oil,muller2019learning} share the same idea of learning from multiple teachers.
Li et al.~\cite{li2018oil} discard bad maneuvers by using a reward based online evaluation of the teachers during training.
Muller et al.~\cite{muller2019learning} use a DNN to fuse multiple controllers and
learn an optimized controller.
Target-driven navigation in static environments is different from the problems above due to the easy acquisition of the optimal expert (the shortest path).
Hence, there is no need to consider the bad demonstrations.
We learn to maximize an expected long-term return provided by environments.
On the other hand, we add an intermediate process to the navigation policy (the generation of the future observation) and predict an action based on the
difference between the current and the future observations.
This makes a more effective and generalizable navigation model.

\textbf{Information Gain-based Approaches.}
Information gain-based strategies have been applied to a variety of robotics problems involving planning and control.
They have been used to study optimal sensor placement and motion coordination for a  target-tracking task~\cite{martinez2006optimal},
derive an information-theoretic metric as a new visual feature for visual servoing~\cite{dame2011mutual},
optimize an information-theoretic objective to improve the informativeness of both local motion primitives and global plans for mapping~\cite{charrow2015information},
facilitate RL to compute good trajectories for scene exploration~\cite{bai2016information},
and generate intrinsic reward to learn an exploration policy~\cite{pathak2017curiosity}.
There are differences in the way mutual information is used in these applications.
However, information gain-based strategies have not been applied to mapless target-driven visual navigation,
the goal of which is to rapidly navigate from a random location in a scene to a specified target.
%Kollar and Roy [24] formulated exploration
%as a constrained optimization problem and used reinforcement
%learning to.
%Bai et al. (2016) propose an method using and show experiments on simplistic map environments.

\textbf{Dynamics Model Learning.}
There is extensive literature on learning a dynamics model, and using this model to train a policy.
Most notable among these is the work from~\cite{racaniere2017imagination} that proposes the Imagination-Augmented Agent, which learns approximate environment models before outputing the action policy.
Pascanu et al.~\cite{pascanu2017learning} propose Imagination-based Planner, which can perform a variable number of imagination steps before any action.
Ha et al.~\cite{ha2018recurrent} incorporate a generative recurrent model into reinforcement learning to predict the future given the past in an unsupervised manner.
However, the goal information is hard-coded in these neural networks and the experimental environments are generally simple and fully observed, leading
to poor generalization to complex, high-dimensional tasks with unseen targets in partially observed scenes.
Pathak et al.~\cite{pathak2018zero} learn an inverse dynamics model based on the demonstrated trajectory way-points from the expert and demonstrate navigation in previously unseen office environments with a TurtleBot.
Although sharing a similar spirit, our work is different from this work.
First, we learn the forward dynamics using a variational generative model, which explicitly models uncertainty over the visual observations, in contrast to the deterministic process in~\cite{pathak2018zero}.
Second, our action policy is directly based on the generated future,
while their predicted action is used for inferring the future.
We design the model with bidirectional information flowing to maximize the mutual information between the action and the adjacent observation pair.

%% file: method.tex
\section{Target-Driven Visual Navigation}
In this section, we begin by outlining the target-driven visual navigation task. We then present our network, which combines an information-theoretic regularization with deep reinforcement learning for this task.

\subsection{Navigation Task Setup}
\label{sec:setup}
We focus on learning a policy for navigating an agent from its current location
to a target in an unknown scene using only visual observations. Our problem is: given a target image $g$, at each time step $t$, the agent receives as input an observation of an environment $x_t$ to predict an action $a_t$ that will navigate the robot to the viewpoint where $g$ is taken.

\textbf{Datasets.} We conduct our experiments on the AI2-THOR, AVD, and some real-world scenarios. AI2-THOR consists of $120$ synthetic scenes in four categories: kitchen, living room, bedroom, and bathroom. Each category includes $30$ scenes, $20$ of which are used for training, $5$ for validating, and $5$ for testing, in accordance with~\cite{wortsman2019}. AVD contains $14$ different households, $8$ of which are used for training, $3$ for validating, and $3$ for testing, as in~\cite{mousavian2019visual}.
We further transfer the learned policies from AVD to some real-world public scenes based on a robotic platform (e.g., TurtleBot); these scenes have never been encountered before.

\textbf{Observations.} In contrast to~\cite{zhu2017}, which stacks four history frames as current inputs at each time step, we utilize four views (RGB images by default) with evenly distributed azimuth angles at each location for current observation $x_t$. The resolution of each view is $300*300$.
%, unless explicitly stated otherwise.

\textbf{Targets.} The navigation target is specified by an RGB image, which contains a goal object. \emph{Please refer to the Appendix for the training and testing goal objects}.
%For each object, AI2-THOR provides all related views, which are used as target images in our training. On AVD, we just use one view for each goal object.
Our model learns to analyze the relationship between the current observation and the target image, and hence we can show generalization to novel targets and scenes that the agent has not previously encountered.

\textbf{Actions.} Each scene in our dataset is discretized into a grid-world navigation graph. The agent acts on these graphs and its action space is determined by the connectivity structures of these graphs as a discrete set: $\mathcal{A}=\{\emph{move forward}; \emph{move back}; \emph{move left}; \emph{move right}; \emph{rotate ccw}; \\ \emph{rotate cw}; \emph{stop}\}$, as defined in~\cite{mousavian2019visual}.
The discrete graph makes it easy to acquire a shortest action path for a target-driven navigation task (e.g., using A-star algorithm).
In this work, we will show how to incorporate the shortest paths during training to learn a navigation controller.

\textbf{Rewards.} Our purpose during policy training is to minimize the length of the trajectory to the navigation targets. Therefore, reaching the target is assigned a high reward value $10.0$ and we penalize each step with a small negative reward $-0.01$. To avoid collision, we design a penalty of $-0.2$ when obstacles are hit during run-time. In addition, we consider the geodesic distance to the goal at each time step, $Geo(x_t; g)$, as in~\cite{savva2019habitat}, and reformulate the reward as:

\vspace{-10pt}
\begin{small}
\begin{equation}
r_t=\left\{
\begin{aligned}
&-0.01 &&\textup{if}~t=0\\
&+10.0 &&\textup{elif}~succeed\\
&-0.2&&\textup{elif}~collide\\
&Geo(x_{t-1},g)-Geo(x_t,g)-0.01&&\textup{otherwise.}
\end{aligned}
\right.
\end{equation}
\end{small}
\textbf{Success measure.} In our setting, the agent runs up to $100$ steps, unless a stop action is issued or a task is successful. The task is considered successful if the agent predicts a stop action,
the goal object is in the field of the current front-view,
and the distance between the current location and the target view location is within a threshold (e.g., $0.5m$ for the AI2-THOR simulator and $1.0m$ for two real-world settings).

\subsection{Information-Theoretic Regularization}
\label{sec:MI}
\begin{figure}[thpb]
\begin{center}
\includegraphics[width=\linewidth]{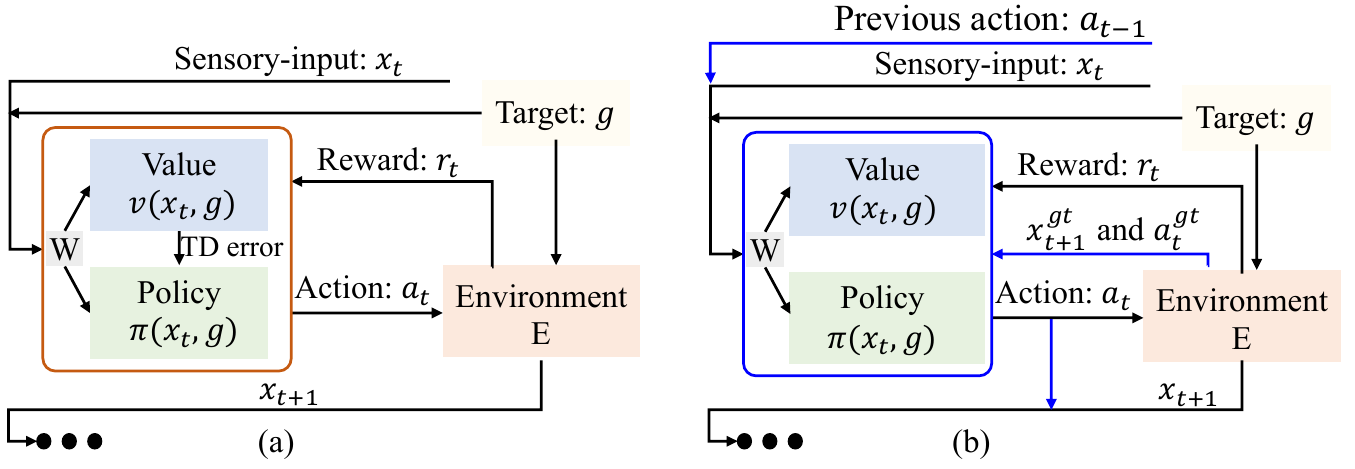}
\end{center}
\caption{Target-driven navigation flow diagram showing how agents interact with the environment. (a) Traditional RL agent  (in the orange square). (b) Our agent (in the blue square). Layer parameters in a gray square are shared by a policy network and a value network. The blue represents the difference between the two flow diagrams. We propose using an information-theoretic regularization to facilitate the traditional RL learning, which requires more information from the environment, e.g., $x_{t+1}^{gt}$ and $a^{gt}_t$.}
\label{fig:Navtask}\vspace{-10pt}
\end{figure}

We formulate the target-driven visual navigation using a deep reinforcement learning framework (TD-A3C). At each time step $t$, the network takes a current observation $x_t$ and a navigation target $g$ as inputs and finally outputs an action distribution $\pi(x_t, g)$ and a scalar $v(x_t, g)$.
We choose action $a_t$ from the policy $\pi(x_t, g)$, and $v(x_t, g)$ is the value of the current policy.
This network can be updated by minimizing a traditional RL navigation loss as in~\cite{zhu2017}. Figure~\ref{fig:Navtask}(a) shows the interaction between the agent and an environment.
However, achieving strong results with one single policy network for all training scenes is difficult, since the agent is very sensitive to the RL reward function and requires extensive training time.
In addition,~\cite{zhu2017} does not consider generalization to previously unseen
environments.
%, which are new houses with different layouts and furniture locations.

In order to address the shortcoming above, we revisit Shannon's mutual information (MI) measure to further reduce the uncertainty in navigation action decisions when a visual observation is given.
Let $x_{t}$ denote the current observation, $x_{t+1}$ denote the next observation, and $a_t$ be the relative action between the two observations. We observe that an agent always abides by a task-independent information-theoretic regularization: there should be high mutual information between the action $a_t$ and the next observation $x_{t+1}$ given the current observation $x_t$. The mutual information $I(x_{t+1}, a_t | x_t)$ is defined as:

\vspace{-10pt}
\begin{small}
\begin{equation}\label{eq:regularization}
\begin{aligned}
&I(a_t, x_{t+1}| x_t)=H(a_t|x_t)-H(a_t|x_{t+1}, x_t)\\
&=\iint p(a_t,x_{t+1}|x_t)\log{p(a_t|x_t,x_{t+1})}da_tdx_{t+1}+H(a_t)\\
&\geq\iint p(x_{t+1}|x_t,a_t)p(a_t|x_t)\log{p(a_t|x_t,x_{t+1})}da_tdx_{t+1}\\
&=\iiint p(x_{t+1}|z)p(z|x_t,a_t)p(a_t)\log{p(a_t|x_t,x_{t+1})}da_tdx_{t+1}dz\\
&=E_{x_{t+1}\sim p(x_{t+1}|z)}[E_{z\sim p(z|x_t,a_t)}[E_{a_t\sim p(a_t)}[\log{p(a_t|x_t,x_{t+1})}]]]
\end{aligned}
\end{equation}
\end{small}
In this setting, we suggest that the action $a_t$ of an agent is unrelated to its current visual observation $x_t$, but in connection with $x_t$ only if the next observation $x_{t+1}$ or a navigation goal $g$ is given. Thus, we have $p(a_t| x_t)$ identically equal to $p(a_t)$, i.e. $H(a_t| x_t)$ is identically equal to $H(a_t)$.
This is different from traditional learning-based visual navigation methods, which tackle individual tasks in isolation, where the goal information is hard-coded in the neural networks and corresponding state descriptions~\cite{zhang2017deep}. Thus, $p(a_t|x_t)\neq p(a_t)$ and these present poor generalization to unexplored targets. In addition, our action space is the deterministic discrete set $\mathcal{A}$, i.e. $a_t\in\mathcal{A}$. Hence, we assume $a_t\sim Cat(1/C)$, where $Cat(\cdot)$ is a categorical distribution and $C$ is the cardinality of $\mathcal{A}$. $H(a_t)\geq 0$ is a constant.
%Further, we introduce a latent variable $z$ as the likes of VAE~\cite{bao2017cvae}, to model the generation of the next observation $x_{t+1}$.
This regularization provides a well-grounded action-observation dynamic model $p(x_{t+1}|x_t,a_t)$ and describes the causality between navigation actions and observational changes $p(a_t|x_t,x_{t+1})$. An agent that seeks to maximize this value will gain a compelling understanding of the dynamics and the causality. This intuition leads us to incorporate the regularization into our navigation learning.

We propose adapting the task-independent regularization above by incorporating some supervision to help learn a strong target-driven visual navigation model, a special case of the lower bound in Equation~\ref{eq:regularization}.
The supervision is from the shortest paths of target-driven navigation tasks. Specifically, at each time step, given the current observation and the target, the optimal next observation $x_{t+1}^{gt}$ and relative action $a^{gt}_t$ are provided as ground truth, see Figure~\ref{fig:Navtask}(b).

To maximize the lower bound, we first assume the next observation $x_{t+1}$ and the ground truth action $a^{gt}_t$ are given and thus the lower bound is converted to a predictive control term as $E_1=E_{a^{gt}_t\sim p(a^{gt}_t)}[\log{p(a_t|x_t,x_{t+1})}]$.
Subsequently, we generate the next observation $x_{t+1}$ by a generative module $z\sim p(z|x_t,a_t), x_{t+1}\sim p(x_{t+1}|z)$, which is most related to the navigation task.
Hence, we use the ground truth action $a_{t}^{gt}$ to guide the generation: $z\sim p(z| x_t, a_{t}^{gt}), x_{t+1}\sim p(x_{t+1}|z)$.
The ground truth $x_{t+1}^{gt}$ is used to help update the generation module though a reconstruction term $E_2=||x^{gt}_{t+1}-x_{t+1}||_2$.
In addition, considering that $a_{t}^{gt}$ is unknown a priori during real navigation and is inherently determined by the navigation target $g$, we design the distribution $q(z|x_t,g)$ to approximate the distribution $p(z|x_t,a_{t}^{gt})$, which formulates a Kullback-Leibler (KL) divergence term as $E_3=\mathcal{K}\mathcal{L}[q(z|x_{t},g)||p(z|a_{t}^{gt},x_{t})]$.
$z\sim p(z|x_t,a_{t}^{gt})$, $x_{t+1}\sim p(x_{t+1}|z)$ and $z\sim q(z|x_t,g)$ constitute our variational auto-encoder module.
Overall, we obtain a variational objective function as:

\vspace{-10pt}
\begin{equation}\label{eq:init_loss}
\begin{aligned}
\mathcal{J}(x_t,g)&=\alpha E_1-\beta E_2-\gamma E_3
\end{aligned}
\end{equation}
The hyper-parameter $(\alpha, \beta, \gamma)$ tunes the relative importance of the three terms: predictive control, reconstruction, and KL.

\subsection{Regularized Navigation Model}
\begin{figure}[thpb]
\begin{center}
\includegraphics[width=\linewidth]{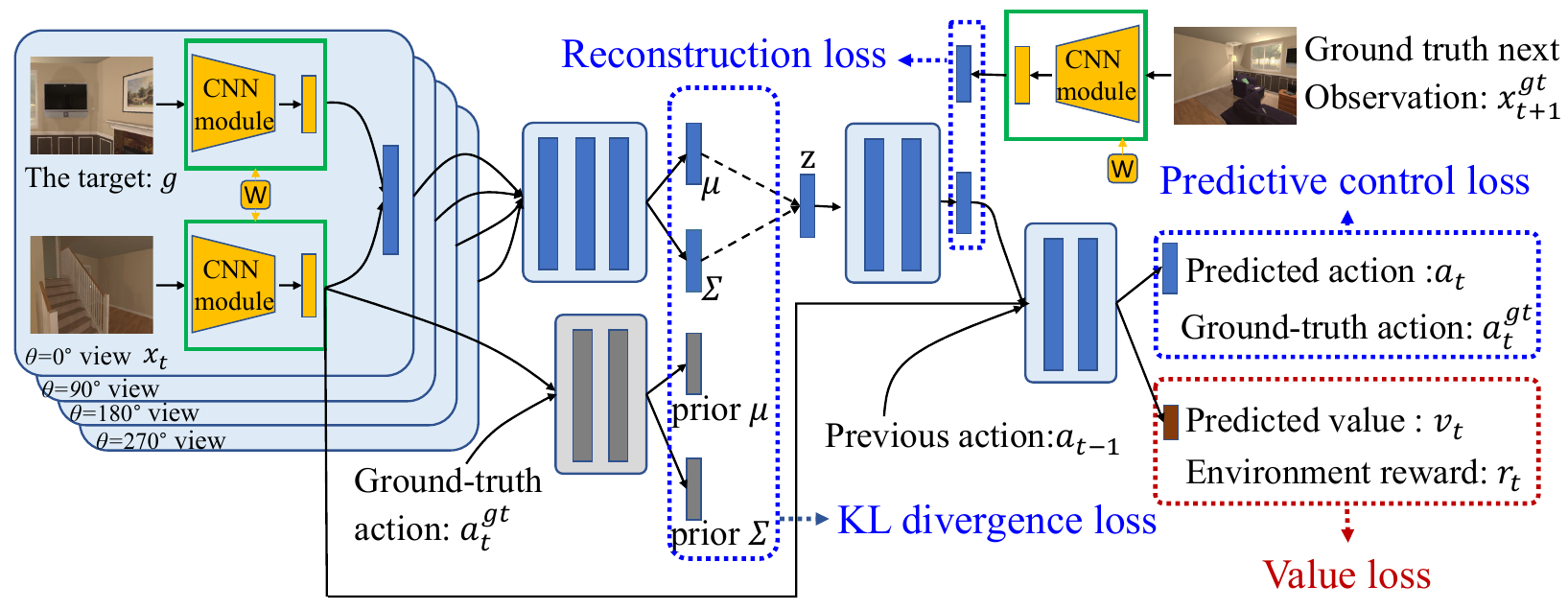}\vspace{-12pt}
\end{center}
   \caption{Model overview. Our model integrates an information-theoretic regularization into an RL framework to constrain the intermediate process of the navigation policy. During training, our network is supervised by the environment reward $r_t$, the shortest path of the task in the form of the ground truth action $a^{gt}_t$, and the ground truth next observation $x^{gt}_{t+1}$. The parameters are updated by four loss terms: the reconstruction, the KL, the predictive control and the value. The first three terms in blue are introduced by the information-theoretic regularization. At test time, the parameters are fixed and our network takes the current observation and the target as inputs to generate the future state. Then it predicts the action based on the future and the current states. Layer parameters in the green squares are shared.
 }
\label{fig:overview}\vspace{-10pt}
\end{figure}

The key idea in reinforcement learning for navigation is finding a policy $\pi(x_t, g)$ that can maximize expected future return.
Within our regularized navigation framework, along with the environment reward, our agent puts a great deal of weight on the ability to understand the action-observation dynamics and the causality between actions and observational changes.
This changes the RL problem to:

\vspace{-10pt}
\begin{equation}
\pi^{\ast}=arg\max_{\pi}E[\sum_{t=0}^{\infty}\tau^t r_{t}+\mathcal{J}(x_t,g)]
\end{equation}
where $r_t$ is a reactive reward provided by the environment at each time step and  $\tau\in(0,1]$ is a discount factor.
Corresponding to the objective, the actor-critic structure in our regularized A3C framework is developed as Figure~\ref{fig:overview}.

\textbf{Policy Network.} The inputs to the policy network are the multi-view images $x_t$ and the target image $g$ at each time step $t$.
The network first learns to reason about some important information from the current observation based on the target, which is then used to generate the next expected observation. This process is supervised by the action-observation dynamics $p(x_{t+1}|x_t,a_t^{gt})$ and the ground truth next observation $x_{t+1}^{gt}$.
Information from the generated observation and the current observation is fused to
form a joint representation, which is passed through the predictive control layer to predict the navigation action.

In addition, we investigate two techniques to improve the training performance.
First, we find that when the previous action $a_{t-1}$ is provided, the agent is less likely to move or rotate back and forth in a scene. This is reasonable since the ground truth action has no chance to contradict the previous action (e.g., move forward vs. move backward). Second, we apply a CNN module $f$ to derive a state representation from an image and hence get the current state $f(x_t)$, the ground truth next state $f(x_{t+1}^{gt})$, and the goal state $f(g)$. \emph{Please refer to the Appendix for the structure.}
We do not directly generate the next observation $x_{t+1}$. We generate the state representation, denoted as $s_{t+1}\sim p(s_{t+1}|z)$ and use this to compute the reconstruction loss and predict the navigation action.
To avoid confusion, we will still use the description of generating the next observation given below.
This simplification reduces the network parameters and the computational cost.
As a result, our  policy is updated by:

\vspace{-10pt}
\begin{equation}\label{eq:obj_func}
\begin{aligned}
\mathcal{L}_p=&\alpha E_{a_{t}^{gt}\thicksim p(a_{t}^{gt})}[-\log{p(a_t|f(x_t),s_{t+1},a_{t-1})}]\\
&+\beta ||s_{t+1}-f(x_{t+1}^{gt})||_2\\
&+\gamma \mathcal{K}\mathcal{L}[q(z|f(x_t),f(g))||p(z|a_{t}^{gt},f(x_t))]
\end{aligned}
\end{equation}

\textbf{Value Network.}
We learn a value function from the penultimate connected layer of our policy $\pi (x_t, g)$, which represents the value of the current policy at the current navigation task, denoted as $v(x_t, g)$.
This is associated with a value loss $\mathcal{L}_v=E_{x_t,r_t}[(R_t-v(x_t, g))^2]$, where $R_t$ is the discounted accumulative reward defined by $R_t=\sum_{i=0}^{T-t}\tau^i r_{t+i}+v(x_{T+1},g)$.
Unlike previous work in~\cite{zhu2017} which directly uses the value $v(x_t,g)$ (embodied as a TD error) to help update the navigation policy, see Figure~\ref{fig:Navtask}(a), our value term $\mathcal{L}_v$ merely affects the shared layers of the policy in Figure~\ref{fig:Navtask}(b).
Hence, our value network functions as an auxiliary task, and we will show that this cooperation consistently outperforms the baseline without it in Section~\ref{sec:Results1}.

Therefore, the overall loss function is $\mathcal{L}=\mathcal{L}_p+\omega\mathcal{L}_v$, where the hyper-parameter is empirically set as ($\alpha=1.0$, $\beta=0.01$,$\gamma=0.0001$, $\omega=0.5$) throughout our experiments.
\emph{Please refer to the supplemental material for more details}.

At test time, three modules $z\sim q(z|x_t, g), x_{t+1}\sim p(x_{t+1}|z)$, and $a_t\sim p(a_t|x_t, x_{t+1}, a_{t-1})$ constitute our controller for the agent to predict the next action given the current observation, the target view, and the previous action.
The controller can navigate robots in unseen scenes, the environment maps (graphs) of which are not known.

%% file: result_discuss.tex
\section{Implementation and Performance}
\label{sec:Implementation and Performance}
Our objective is to improve the cross-target and cross-scene generalization of target-driven navigation. In this section, we evaluate our model compared to baselines based on standard deep RL models and$/$or traditional imitation learning.
We also provide ablation results to gain insight into how performance is affected by changing the structures.

\subsection{Baselines and Ablations}
We compare our method with the following visual navigation models:
(1) \textbf{TD-A3C} is the target-driven visual navigation model from~\cite{zhu2017} and is trained using standard reinforcement learning.
This was originally designed for scene-specific navigation and it is difficult to achieve strong results with one single policy network for all training scenes.
We assist the policy learning by using the reward function and previous action as ours.
(2) \textbf{TD-A3C(BC)} is a variation of the TD-A3C. It is trained using behavioral cloning (BC). Both the CNN module and the input are the same as ours.
The main difference from our method is how the supervision is exploited.
(3) \textbf{Gated-LSTM-A3C(BC)} is an LSTM-based variant of A3C model adapted from~\cite{wu2018building}, which is trained with BC and provided with the previous action. The goal is specified as an image, and the model is also provided with the same multi-view images as in ours.
(4) \textbf{GSP} is a goal-conditioned skill policy in~\cite{pathak2018zero},
which generates the next observation as an auxiliary task rather than using the generation for navigation control. We reimplement the work\footnote{https://github.com/pathak22/zeroshot-imitation} and train it on our datasets.
(5) \textbf{SAVN} proposes a self-adaptive visual navigation model~\cite{wortsman2019}, which shows strong results on novel scene adaption on AI2-THOR. It does not, however, support adaptation to novel targets.
(6) \textbf{TD-Semantic} is a navigation model from~\cite{mousavian2019visual}. The method predicts the cost of an action, which is supervised by shortest paths of navigation tasks.
(7) \textbf{NeoNav} is our previous work~\cite{wuneonav}.
(8) \textbf{Ours-NoVal} is a variant of our method, which does not consider the value of our
navigation policy.
(9) \textbf{Ours-FroView} is a variant of our method and takes the current front-view to generate the future observation.
(10) \textbf{Ours-NoGen} is a variation of our model that predicts $x_{t+1}$ directly from the current observation $x_t$ and the target $g$ without a stochastic latent space.
(11) \textbf{Ours-VallinaGen} is  a variant of ours, in which the latent space $z\sim q(z|x_t,g)$ is constrained by the standard normal distribution prior $p(z)$.

We train and evaluate these models on the datasets described in Section~\ref{sec:setup}.
Except for TD-A3C and SAVN, all alternatives are trained with supervision from shortest paths of navigation tasks, although they use different methods, e.g., behavioral cloning as TD-A3C(BC), setting waypoints from experts as GSP, predicting navigation action cost as TD-Semantic, generating next expected observation as NeoNav, etc.
We incorporate supervision into RL frameworks for mapless visual navigation, since RL for high-dimensional observations empirically suffers from sample inefficiency~\cite{Lukasz2020}.
We evaluate these models on two metrics, success rate (SR) and success weighted by (normalized inverse) path length (SPL), as defined in~\cite{yang2018visual}.
%\emph{More details are provided in~\cite{wu2019reinforcement}.}

\vspace{-8pt}
\begin{table}[h]
\centering
\caption{Average navigation performance (SR and SPL in \%) comparisons on unseen scenes from AI2-THOR.
\label{tab:table1}}\vspace{-10pt}
\scalebox{0.90}{\begin{tabular}{c|c|c|c|c|c}
\cline{1-6}
\hline
Evaluations& Models &\multicolumn{2}{ |c }{All}&\multicolumn{2}{ |c }{$L\geq 5$} \\\cline{3-6}
  &  &SR &SPL &SR &SPL\\
\hline
 &Random &  1.2 & 0.7& 0.6 &  0.3  	\\\cline{2-6}
&TD-A3C~\cite{zhu2017}&  20.0 & 4.0 & 12.9 &  2.6 \\\cline{2-6}
&TD-A3C~\cite{zhu2017}(BC)&  23.0 & 7.9 & 13.4 &  3.7 \\\cline{2-6}
Unseen,&Gated-LSTM-A3C~\cite{wu2018building}(BC)&  29.1 & 10.5 & 19.2 &  5.1 \\\cline{2-6}
scenes&GSP~\cite{pathak2018zero}&  34.4 & 12.5 & 27.9 & 8.3  \\\cline{2-6}
Known&NeoNav~\cite{wuneonav}&  30.2 & 11.9 & 23.6 & 10.1\\\cline{2-6}
targets&Ours&  \textbf{45.7} & \textbf{25.8} & \textbf{41.9} & \textbf{ 24.8}  \\\cline{2-6}
P=17.7\%&Ours-NoVal&  34.9 & 18.1 & 27.3 & 14.5\\\cline{2-6}
&Ours-FroView&  32.3 & 10.3 & 29.8 & 9.4 \\\cline{2-6}
&Ours-NoGen&  41.2 & 23.8 & 38.5 &  22.2\\\cline{2-6}
&Ours-VallinaGen&  37.5 & 17.7 & 34.0 &  15.9 \\\cline{2-6}
\hline
\hline
 &Random &  2.0 & 1.0& 0.6 &  0.4 \\\cline{2-6}
&TD-A3C~\cite{zhu2017}&  10.1 & 1.9 & 6.3 & 1.1\\\cline{2-6}
&TD-A3C~\cite{zhu2017}(BC)&  12.3 & 2.4 & 7.5 &  1.6 \\\cline{2-6}
Unseen&Gated-LSTM-A3C~\cite{wu2018building}(BC)&  30.0 &11.4 & 26.7 &  8.6 \\\cline{2-6}
scenes,&GSP~\cite{pathak2018zero}&  27.5 & 8.3 & 23.4 & 6.7  \\\cline{2-6}
Novel&NeoNav~\cite{wuneonav}&  27.4 & 13.1 & 22.1 & 9.3 \\\cline{2-6}
targets&Ours&  \textbf{37.7} & \textbf{20.5} & \textbf{35.4} &  \textbf{19.7}  \\\cline{2-6}
P=16.0\%&Ours-NoVal&  31.6 & 10.3 & 28.9 & 9.4 \\\cline{2-6}
&Ours-FroView&  24.6 & 7.8 & 23.0 & 6.9 \\\cline{2-6}
&Ours-NoGen&  35.7 & 19.1 & 31.6 &  17.4 \\\cline{2-6}
&Ours-VallinaGen&  31.4 & 13.9 & 29.4 &  12.7 \\\cline{2-6}
\hline
\end{tabular}}\vspace{-12pt}
\end{table}

%\subsection{Evaluation Metrics}
%We evaluate these models on two metrics, success rate and success weighted by (normalized inverse) path length (SPL) [xx] (Anderson et al. 2018). Success rate is the fraction of the runs that successfully navigate to the goal. SPL considers both success rate and the path length traveled:
%, where N is the number of navigation tasks, Si is a binary indicator of success in the i-th task. Pi and Li denote the actual path length and the shortest path distance for the i-th task, respectively.

\subsection{Results on the AI2-THOR}
\label{sec:Results1}
\textbf{Generalization.} We analyze the cross-target and cross-scene generalization ability of these models on AI2-THOR.
The evaluation is divided into two different levels on our testing set, $\{\emph{Unseen}~\emph{scenes},~\emph{Known}~\emph{targets}\}$ and $\{\emph{Unseen}~\emph{scenes},~\emph{Novel}~\emph{targets}\}$.
Each level of evaluation contains $1000$ different navigation tasks.
~\cite{savva2019habitat} proposes using the ratio of the shortest path distance to the Euclidean distance between the start and goal positions to benchmark navigation task difficulty.
In each evaluation, we compute the percentage $P$ of the tasks that have a ratio within the range of $[1, 1.1]$ and evaluate the performance on all tasks and on tasks where the optimal path length $L$ is at least $5$.

Table~\ref{tab:table1} summarizes the results. First, we observe a higher generalization performance for the model with supervision by comparing the results from TD-A3C and TD-A3C(BC). We believe that it is more challenging for RL networks to discover the optimal outputs in the higher-order control tasks.
%In addition, pretraining on ImageNet (TD-A3C) does not offer better generalization, since the features required for ImageNet are different from those needed for navigation.
Subsequently, considering the navigation performance difference between TD-A3C(BC) and ours, we see that the idea of generating the future before acting and acting based on the visual difference, works better than directly learning a mapping from raw images to a navigation action.
We also compare ours with Gated-LSTM-A3C(BC), which uses an LSTM-based memory and has access to shortest paths during training as in our method.
Our model can consistently outperform the LSTM-based baseline.
The model, GSP, also trained with an inverse dynamics model, acquires relatively
better navigation performance compared to TD-A3C(BC) and Gated-LSTM-A3C(BC).
However, our model gets better results compared with the baseline, which indicates the proposed information-theoretic regularization brings us better generalization for unseen scenes and novel objects.
In addition, our method shows at least $10.3\%$ improvement in average SR and $7.4\%$ improvement in average SPL over NeoNav for both cross-target and cross-scene evaluation.
The results demonstrate that our regularized navigation model possesses more advantages in target-driven visual navigation.

To further compare ours with SAVN~\cite{wortsman2019}, we adapt our navigation model to be driven by a target object class, which is given as a Glove embedding, as in SAVN.
The experiment is conducted on AI2-THOR with the same training/testing split, evaluation navigation tasks, and success criterion as~\cite{wortsman2019}.
Our model outperforms SAVN with navigation performance (SR and SPL in $\%$) increased by $3.9/7.1$ for all navigation trajectories ($45.8/23.8~vs.~41.9/16.7$) and $13.8/8.2$ for trajectories of at least length $5$ ($43.1/22.4~vs.~29.3/14.2$).
Our information-theoretic regularization supports both
more effective and more efficient navigation.

\begin{figure}[thpb]
\begin{center}
\includegraphics[width=\linewidth]{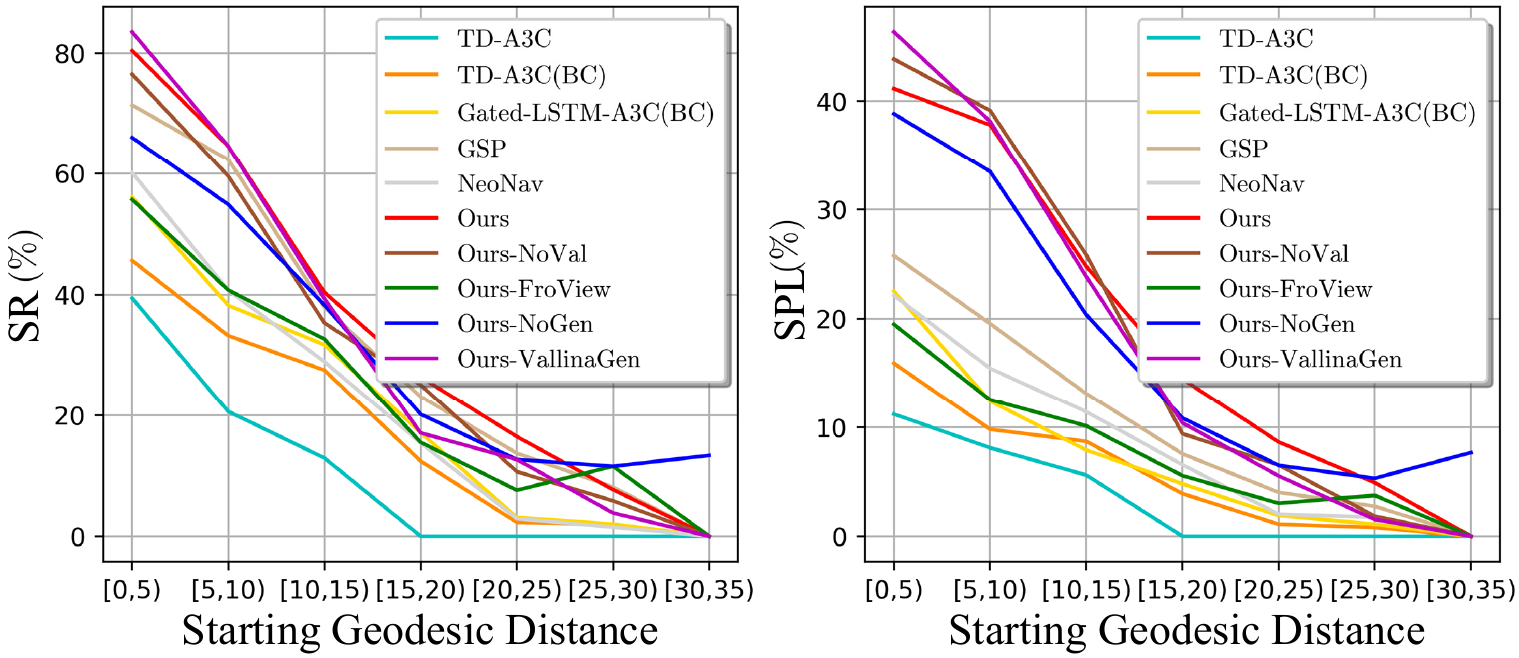}
\end{center}\vspace{-10pt}
   \caption{We report SR and SPL performance as a function of starting geodesic distance from the target.
 }\vspace{-12pt}
\label{fig:geo}
\end{figure}

\textbf{Ablation.} The ablation on different inputs (Front-view vs. Multi-view) demonstrates that it is easier to generate the next observation when the current information is rich.
We also conduct the ablation  with four history frames as current inputs, which is difficult to converge during training.
We consider that there is no direct connection between the random history and the next observation, which is most related to the current observation and the target.
Hence, it is reasonable to generate the next observation from the current multi-view observation rather than from the history frames.

It is difficult for a navigation agent to learn to issue a stop action at a correct location,
since there is only one situation with stop action but many cases with other actions during a navigation task, leading to training data imbalance~\cite{zhao2018triangle}.
We first show the cross-target and cross-scene navigation results in a simpler case where the stop signal is provided by the environment rather than being predicted by an agent.
The navigation performances (SR and SPL in $\%$) of our model and Ours-NoVal are: $41.6/27.2$ and $38.7/19.4$, respectively.
As expected, both models demonstrate higher performances than the performances in Table~\ref{tab:table1} ($37.7/20.5$ and $31.6/10.3$, respectively).
Additionally, we see the performance gap of our model is much smaller than Ours-NoVal, meaning we handle the data imbalance better.
We consider that the value prediction is critical in learning to issue a stop action.
The different stages in navigation tasks can be distinguished by their discounted accumulative reward in RL, and the stage close to the target with a large accumulative reward updates the policy more, which eases the data imbalance.

Based on the ablation on the generation process, we conclude that learning a stochastic latent space is often more generalizable than learning a deterministic one (Ours vs. Ours-NoGen), since the former explicitly models the uncertainty over visual images. However, when the latent space is over-regularized by the standard normal distribution prior, the situation is worse (Ours-VallinaGen vs. Ours).

%\begin{figure}
%\begin{center}
%\includegraphics[width=\linewidth]{./fig_dis_colli.pdf}
%\end{center}\vspace{-5pt}
%   \caption{Analysis. (a) We fit the ending geodesic distance distribution of all learning models over $1000$ navigation tasks. The Init represents the starting geodesic distance distribution of these tasks. (b) We report the collision action percentages of all learning models as the navigation proceeds.
% }\vspace{-12pt}
%\label{fig:dis}
%\end{figure}

\textbf{Geodesic distance.} We further analyze the navigation performance as a function of the geodesic distance between the start and the target locations in Figure~\ref{fig:geo}.
This is based on the $1000$ tasks from the $\{\emph{Unseen}~\emph{scenes},~\emph{Known}~\emph{targets}\}$ evaluation.
As can be seen, the geodesic distance is highly correlated with the difficulty of navigation tasks and the performance of all methods degrades as the distance between the start and the target increases.
Our model outperforms all alternatives in most cases.

\begin{figure}[thpb]
\begin{center}
\includegraphics[width=\linewidth]{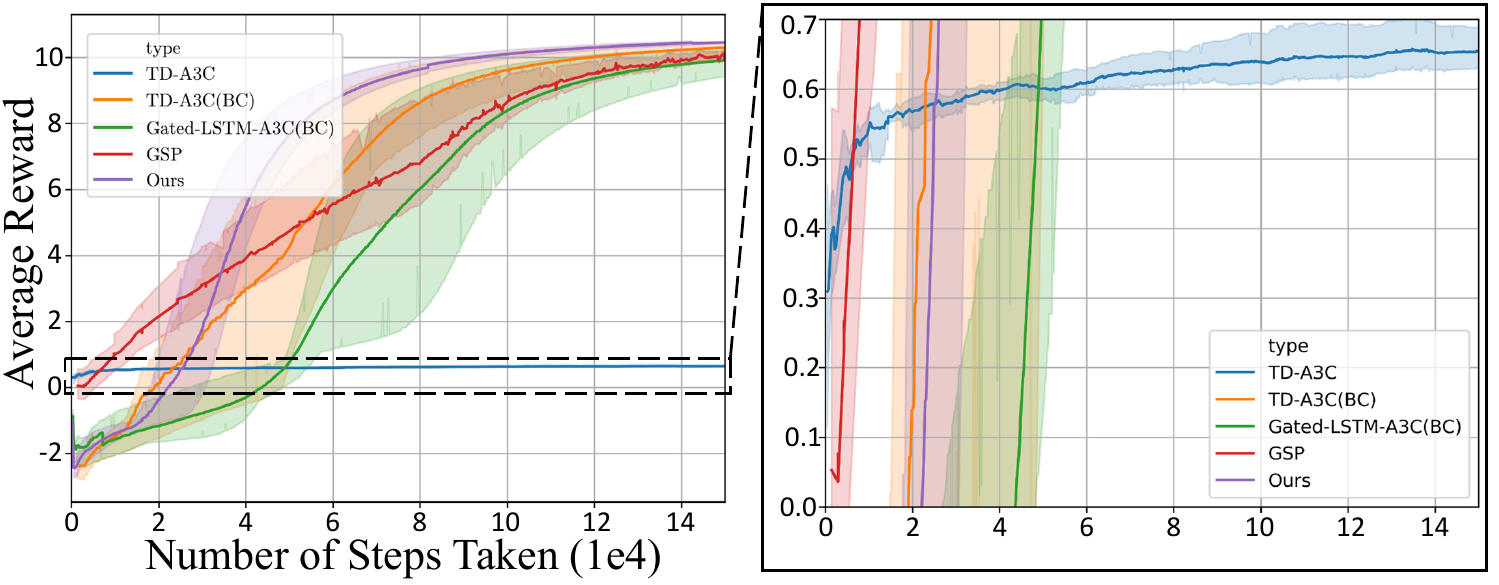}
\end{center}\vspace{-10pt}
   \caption{Average reward curves (left) on AVD during training with a zoomed-in view (right).
   Our model outperforms these alternative navigation methods in terms of learning speed and the final average reward.
 }\vspace{-12pt}
\label{fig:reward}
\end{figure}

\subsection{Results on the AVD}
To evaluate the generalization ability in the real world, we train and evaluate our model and some alternatives based on the training and testing splits on AVD.

\textbf{Navigation driven by target images.}
We first present the navigation results driven by target images.
Figure~\ref{fig:reward} shows the average returns during training from TD-A3C, TD-A3C(BC), Gated-LSTM-A3C(BC), GSP, and our model. We train five different instances of each algorithm with different random seeds, with each performing one evaluation every $200$ navigation episodes.  We plot the average return curves with error bands representing the standard deviation. The results show that
navigation models with supervision (e.g., TD-A3C(BC), Gated-LSTM-A3C(BC), GSP) learn considerably faster than a pure RL-based navigation model (e.g., TD-A3C).
Our proposed model outperforms these baselines both in terms of sample efficiency and the final performance.
In Table~\ref{tab:diffscene}, we also report the average values of success rate and SPL with standard deviations of $1000$ navigation tasks ($P=15.0\%$) randomly sampled from unseen scenes in AVD.
We observe that all five learning models demonstrate average performance decreases compared to the results on AI2-THOR in Table~\ref{tab:table1}, since both the training scenes and the target views in AVD are limited and the real-world scenes are more complex in structure than synthetic scenes.
In addition, our model has relatively larger variance than TD-A3C(BC) and GSP while retaining a better navigation performance than other listed baselines.
We consider that our intermediate generative process increases the stochasticity of navigation control, but the proposed information-theoretic regularization generally brings more useful information for navigation tasks,
which is critical for policy learning from the perceptible environment.

\begin{figure}[thpb]
\begin{center}
\includegraphics[width=\linewidth]{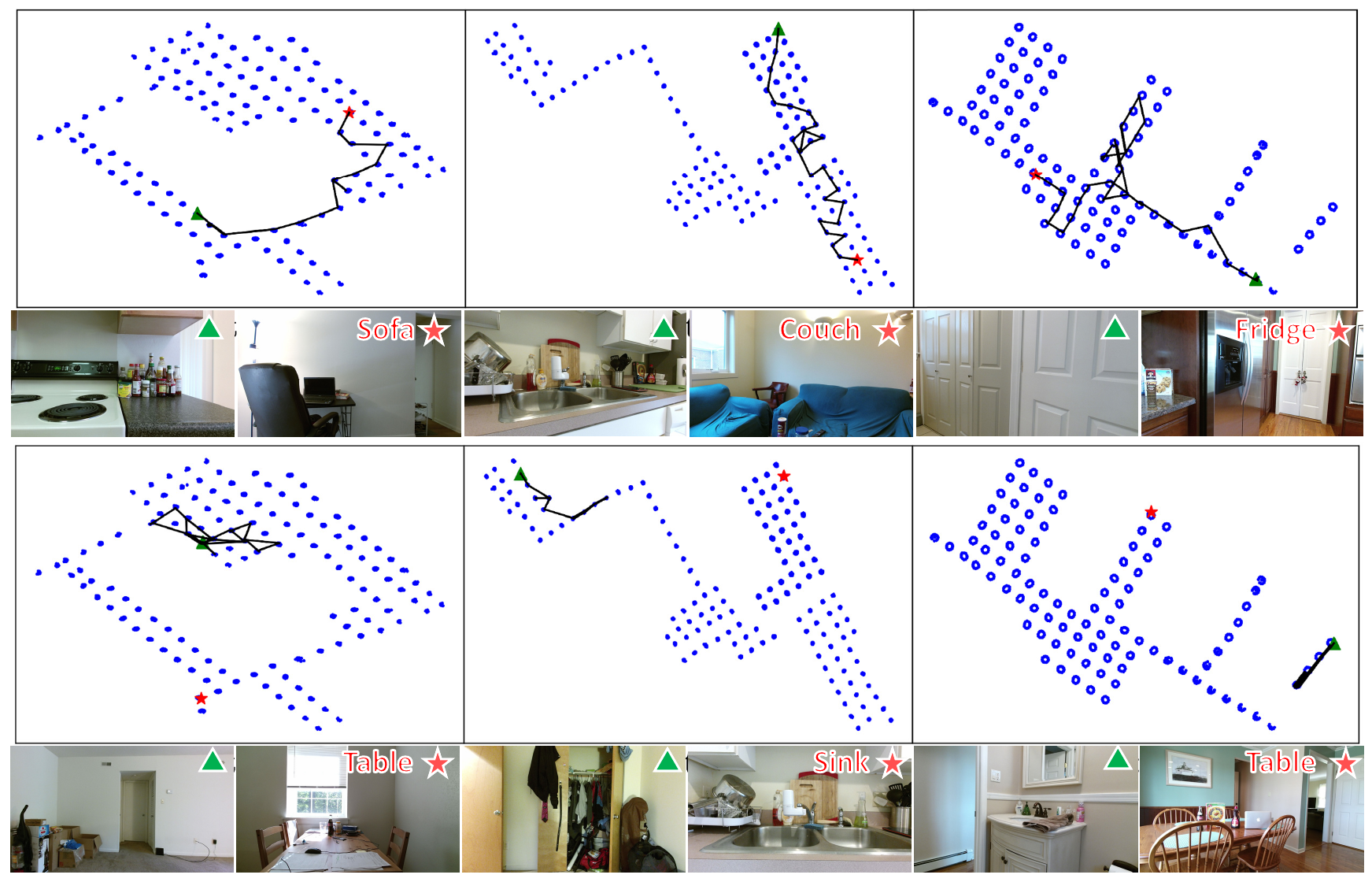}
\end{center}\vspace{-8pt}
   \caption{Visualization of some typical success and failure cases of our method from AVD. The blue dots represent reachable locations in the scene. Green triangles and red stars denote starting and goal points, respectively.
 }\vspace{-12pt}
\label{fig:path}
\end{figure}

We visualize six navigation trajectories from our model in Figure~\ref{fig:path}. These tasks are all characterized by unknown scenes, and long distances between the start points and the targets. For the tasks in the first row, our agent can navigate to the targets successfully, but for the last three tasks, our model fails to finish within the maximum steps.
The problems include thrashing around in space without making progress (see the
first and third trajectories in the second row), getting stuck in the corridor (see
the second trajectory in the second row), and navigating around tight spaces
(e.g, the bathroom where the fourth trajectory starts).

\begin{table}[h]
\centering
\caption{Average navigation performance (SR, SPL and CR in \%) comparisons on unseen scenes from AVD and realistic scenarios (RS).
\label{tab:diffscene}}\vspace{-10pt}\scalebox{0.90}
{\begin{tabular}{c|c|c|c|c}
\cline{1-5}
\hline
&\multicolumn{2}{ |c| }{AVD}& \multicolumn{2}{ |c }{RS} \\\cline{2-5}
Model  &SR &SPL &SR&CR\\
\hline
Random &2.8$_{(0.9)}$ & 1.8$_{(0.4)}$&2.0&62.0\\\cline{1-5}
TD-A3C~\cite{zhu2017}&   9.3$_{(2.4)}$ & 2.9$_{(1.1)}$&2.0&58.0\\\cline{1-5}
TD-A3C~\cite{zhu2017}(BC)&   15.9$_{(1.9)}$ & 6.1$_{(0.9)}$&8.0&56.0\\\cline{1-5}
Gated-LSTM-A3C~\cite{wu2018building}(BC)&  13.3$_{(2.7)}$ & 5.8$_{(1.6)}$&10.0&46.0\\\cline{1-5}
GSP~\cite{pathak2018zero}& 19.3$_{(1.1)}$ & 5.5$_{(0.6)}$ &24.0&48.0\\\cline{1-5}
Ours&  \textbf{23.1}$_{(2.1)}$ & \textbf{13.5}$_{(1.0)}$&28.0&40.0\\\cline{1-5}
\hline
\end{tabular}}\vspace{-12pt}
\end{table}
\textbf{Navigation driven by target labels.}
We also adapt our method to compare it with TD-Semantic~\cite{mousavian2019visual}, in which the navigation goal is defined in the form of a one-hot vector over a prescribed set of semantic labels, $\{\emph{Couch},~\emph{Table},~\emph{Refrigerator},~\emph{Microwave},~\emph{TV}\}$.
The experiment is conducted on AVD with the same training/testing split,  evaluation tasks, and success criterion as~\cite{mousavian2019visual}.
%TD-Semantic provides several ablations on different input modalities, including RGB, depth and semantic information.
While sharing the same idea of improving the training by using the supervision from shortest paths of navigation tasks, our method outperforms TD-Semantic by $22\%$ for RGB input ($53\%~vs.~31\%$), and $28\%$ for depth input ($59\%~vs.~31\%$) for average success rate on the AVD testing set. Our method (with depth input) shows a $5.4\%$ improvement in average success rate compared to TD-Semantic with semantic input ($59.0\%~vs.~53.6\%$), which is provided by some state-of-the-art detectors and segmentors. The best performances of two methods over various target labels are presented in the Appendix.
We suggest that our information-theoretic regularization helps learn a controller that can analyze the relationship between visual observation and the target and extract some important information to guide navigation.
This process is not affected by the target format, e.g., a semantic label or a view image.
%The best performances of two methods over various target labels are presented in Table~\ref{tab:semg}.
\subsection{Results on the Real World}
Moving to the real-world scenarios further shows the generalization capabilities of the proposed navigation models and the robustness against indoor layouts, robot types and sensor types. The models are trained based purely on the discrete dataset (e.g., AVD), and the real-world environments are continuous and unknown to the agents.

\begin{figure}[thpb]
\begin{center}
\includegraphics[width=\linewidth]{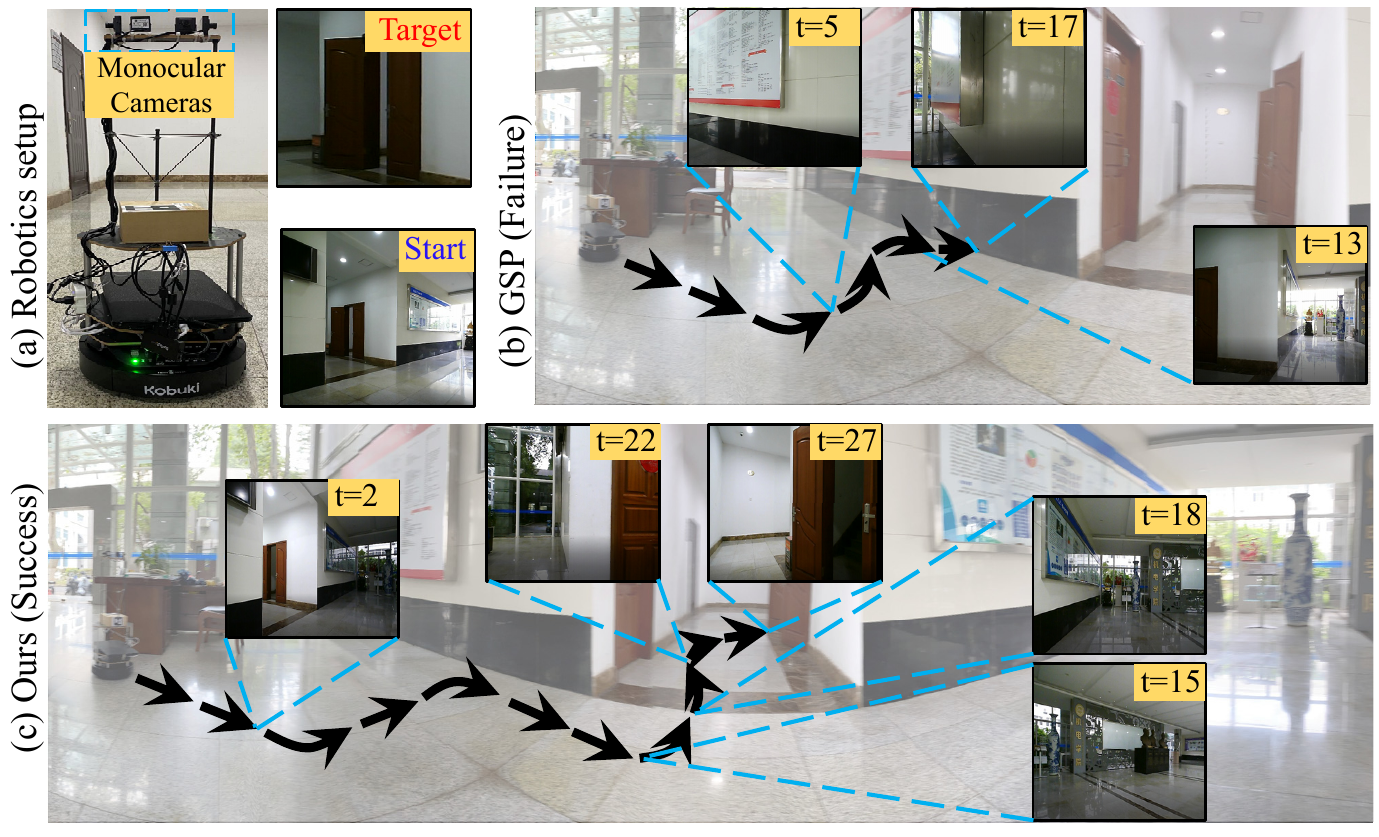}
\end{center}\vspace{-10pt}
   \caption{Qualitative examples. (a) The robotics system setup and the navigation task. (b) The trajectory based on the GSP baseline. (c) The trajectory based on our model.
 }\vspace{-12pt}
\label{fig:realnav}
\end{figure}

\textbf{Robotic setup.} We demonstrate the proposed model using a TurtleBot. The configuration of the TurtleBot is shown in Figure~\ref{fig:realnav}(a), which consists of a differential wheeled moving base Kobuki and four RGB Monocular cameras equipped at the top of the robot.
The proposed system takes as input data from four real-time camera sensors and a target image at each time step, to predict the optimal navigation action.
The action command is converted to the wheel velocity and passed to the robot.
For example, the \emph{move right} action in $\mathcal{A}$ is converted to rotate right at $45^\circ/s$ for $2s$, move forward at $0.25m/s$ for $2s$, and rotate left at $45^\circ/s$ for $2s$. These commands are published with a frequency of $5$Hz.
It is complex due to the movement direction restrictions of the TurtleBot.
%For on-board computations we resort to an Intel NUC with an $i7-5557U$ processor and without any GPU, running Ubuntu $16.04$ and ROS [xx] as a middleware.

\textbf{Transfer to the real world.} Experiments were conducted on a floor (approx. $400m^2$) of an academic building. We evaluate the robot with $50$ randomly sampled navigation tasks in the scene.
A navigation task is regarded as a success only if the robot stops near the target (e.g., $1$m) within $100$ steps, and we consider it a failure if the robot collides with an obstacle or does not reach the goal within the step limit.
Although the model is trained on the discrete household dataset, it can transfer to the realistic public scenarios and exhibit robustness towards random starting points, varying step lengths, changes in illumination, target types, scene layouts, etc.
A quantitative analysis of these navigation tasks is provided in Table~\ref{tab:diffscene} (the right-most column), where the average success rates and collision rates (CR, the rate of collision cases to all navigation cases) are listed.
%While xx does achieve the same collision avoidance capabilities as R-IL,
We observe that the existence of the proposed information-theoretic regularization enables better transfer to new environments. However, all models present consistently high collision rates during navigation, since realistic evaluation, characterized by continuous space and robotic movement deviation, is very challenging.
Extension to depth input or simultaneous mapping~\cite{chaplot2020object} would make the method applicable in more general scenarios.
Figure~\ref{fig:realnav} qualitatively compares our method with the GSP baseline.
The baseline generally gets stuck behind the obstacle (e.g. the wall) and tries hard to move forward, while our method finds the way towards the door and issues the stop action close to the target.
\emph{Examples are provided in the supplementary video.}

%% file: conclusion.tex
\section{Conclusion}
We propose integrating an information-theoretic regularization into a deep reinforcement learning framework for the target-driven task of visual navigation.
The regularization maximizes the mutual information between navigation actions and visual observations, which essentially models the action-observation dynamics and
the causality between navigation actions and observational changes.
By adapting the regularization for target-driven navigation, our agent further learns to build the correlation between the observation and the target.
The experiments on the simulation and the real-world dataset show that our model outperforms some baselines by a large margin in both the cross-scene and the cross-target navigation generalization.
Experiments using the TurtleBot robot demonstrate the transfer capability of the proposed navigation model, which is easy to deploy on-robot.

In this work, training requires supervision from expert trajectories, which are generated based on the topology graphs of training scenes. For future work, we will investigate how real-world human demonstrations can be leveraged and how the model can be extended to dynamic environments. In addition, extending some state-of-the-art model-free and model-based deep RL algorithms to the target-driven visual navigation problem for exploring better navigation generalization is also a great topic for future work.

%% file: appendix.tex
\section{Appendix}
\subsection{Navigation Targets}

\begin{table*}
\centering
\caption{Training and testing split of object classes over scene categories of AI2-THOR.}
\label{tab:navtarget}\vspace{-6pt}
\resizebox{.80\textwidth}{!}{
\begin{tabular}{c|c|c}
\cline{1-3}
\hline
Room type & Train objects & Test objects\\\cline{1-3}
\hline
\hline
Kitchen &  Toaster, Microwave, Fridge, CoffeeMaker, GarbageCan, Box,&StoveBurner, Cabinet,\\
&Bowl, Apple, Chair, DiningTable, Plate, Sink,SinkBasin &HousePlant\\\cline{1-3}
Living room &  Pillow, Laptop, Television, GarbageCan, Box, Bowl,&Statue, TableTop\\
&Book, FloorLamp, Sofa&HousePlant\\\cline{1-3}
Bedroom&  Lamp, Book, AlarmClock, Bed, Mirror& Cabinet, Statue\\
&Pillow, GarbageCan, TissureBox, &Dresser, LightSwitch\\\cline{1-3}
Bathroom &  Sink, ToiletPaper, SoapBottle, LightSwitch, Candle, &Cabinet, Towel\\
&GarbageCan, SinkBasin, ScrubBrush& TowelHolder\\\cline{1-3}
\end{tabular}}\vspace{-6pt}
\end{table*}

Our navigation targets are specified by images, which contain goal objects, such as dining tables, refrigerators, sofas, televisions, chairs, etc.
AI2-THOR~\cite{zhu2017} provides all visible RGB views for each goal object. These views are collected based on three conditions. First, the view should be from the camera's viewport. Second, the goal object should be within a threshold of distance from the agent's center ($1.5m$ by default).
Third, a ray emitted from the camera should hit the object without first hitting another obstruction. In our experiments on AI2-THOR, we have access to about $18231$ different target views from $80$ training scenes for training.
In Table~\ref{tab:navtarget}, we provide the split of object classes used in the training and testing processes of all learning models.
For AVD~\cite{ammirato2017dataset}, we manually select $120$ target views in depth from the training split ($8$ scenes), including some common objects as $\{\emph{Couch}, \emph{Table}, \emph{Fridge}, \emph{Microwave}, \emph{Sink}, \emph{TV}, \emph{Cabinet}, \emph{Toaster}, \\ \emph{GarbageCan}, \emph{Door}, \emph{Chair}, \emph{Bed}, \emph{Dresser}, \emph{Mirror}\}$.

\begin{figure*}[thpb]
\begin{center}\vspace{-5pt}
\includegraphics[width=0.85\textwidth]{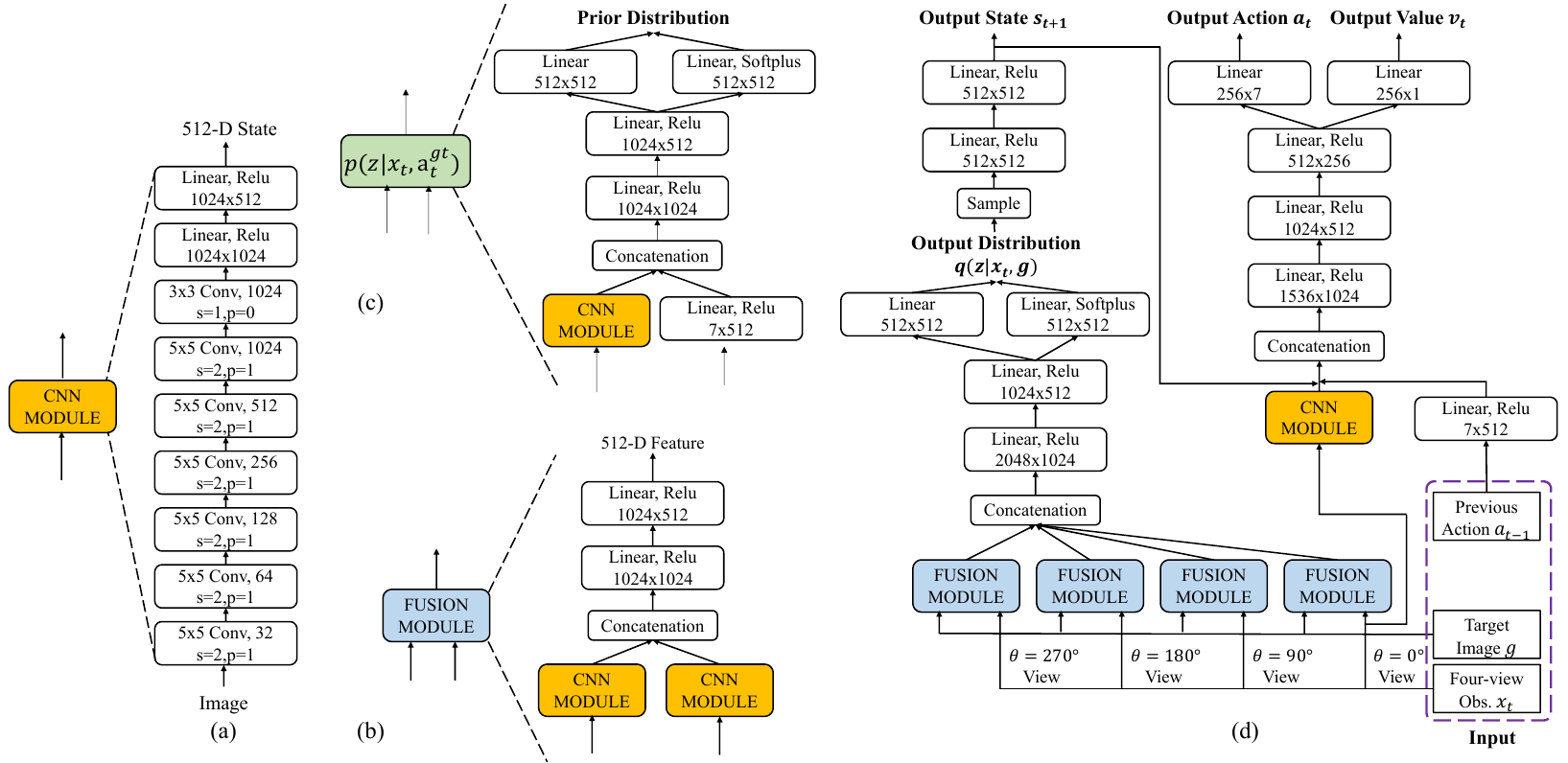}
\end{center}
   \caption{Model architecture. The overview is given (d) with blowups
of (a) the CNN module (the orange portion), (b) the fusion module (the blue portion) and (d) the prior distribution (the green portion).}\vspace{-12pt}
\label{fig:cnn}
\end{figure*}

\subsection{Network Architecture}
Our CNN module for deriving a state representation from an image is presented in Figure~\ref{fig:cnn}(a). By default, spectral normalization is used for the first six layers, which can prevent the escalation of parameter magnitudes and avoid unusual gradients~\cite{miyato2018spectral,zhang2018self}.
The activation function used is LeakyReLU $(0.1)$.
At each time step $t$, we take the four-view observation $x_t$ as well as the target $g$ as inputs and extract a $512$-D state vector for each of them.
We concatenate each view state with the target state to get a fused feature (see Figure~\ref{fig:cnn}(b)).
In Figure~\ref{fig:cnn}(d), four feature vectors are then used to infer a vector of latent variables of dimension $512$ with a MLP. Here, a KL divergence loss is minimized to impose the distribution of the latent variables to match a prior distribution $p(z|x_t, a_t^{gt})$ from Figure~\ref{fig:cnn}(c), which is estimated from the current observation $x_t$ (front view only) and the ground-truth action $a_t^{gt}$.
The latent vector $z\sim q(z|x_t,g)$ is used to generate a state $s_{t+1}$ of next observation, which is under the supervision of ground truth next observation $x_{t+1}^{gt}$.
Subsequently, the generated state $s_{t+1}$ of next observation ($512$-D), the state of front view observation ($512$-D), and the feature ($512$-D) extracted from the
previous action $a_{t-1}$ ($7$-D one-hot vector) are combined together to predict the navigation action $a_t$ ($7$-D) and get the evaluation value $v_t$ ($1$-D).
The ground-truth action $a_t^{gt}$ and environment reward $r_t$ are used to help update this module.

\subsection{Implementation Details}
We train our model using $6$ asynchronous workers and then back-propagate through time for every $10$ unrolled time steps.
The batch size is $60$ for each back-propagation. We use RMSprop optimizer~\cite{tieleman2012divide} to update the network parameters with a learning rate of $1e^{-4}$ and a smoothing constant of $0.99$.
Our model is trained and tested on a PC with $12$ Intel(R) Xeon(R) W-$2133$ CPU, $3.60$ GHz and a Geforce GTX $1080$ Ti GPU.
The training configurations of our ablation models and alternatives, including TD-A3C, TD-A3C(BC), Gated-LSTM-A3C(BC), are much the same as ours.
For all compared models, training on AI2-THOR is carried out in four stages, starting with $20$ kitchens, to gradually increase by $20$ scenes (namely, a scene category) at each next stage. This ensures fast convergence in training scenes.
Training on $8$ scenes from AVD for all learning models is continuous.
The training time of our model is about $60$ hours on $8$ scenes of AVD and $120$ hours on $80$ scenes of AI2-THOR.
We take the model for evaluation which performs best on the validation set.

\begin{table}
\centering
\caption{Comparing navigation performance (SR and SPL in \%) on different scene categories on AI2-THOR with stop action.}\vspace{-6pt}
\label{tab:table2}
\scalebox{0.95}{
\setlength{\tabcolsep}{0.5mm}
{\begin{tabular}{l|c|c|c|c}
\cline{1-5}
\hline
Category & Kitchen & Living room & Bedroom  & Bathroom \\
& P=15.2\%& P=15.6\%& P=20.0\%&P=20.0\%\\
\hline
\hline
Random&  0.0 / 0.0 &  1.6 / 1.0&  2.0 / 1.1&  1.2 / 0.7 \\\hline
TD-A3C~\cite{zhu2017} &  17.4 / 3.1 &  13.2 / 2.1 &  16.9 / 1.9 &  32.4 / 9.0 \\\hline
TD-A3C~\cite{zhu2017}(BC) &  21.3 / 7.8 & 18.2 / 5.2  &22.4 / 8.1  & 30.1 / 10.4\\\hline
Gated-LSTM-A3C~\cite{wu2018building}(BC) &  28.2 / 10.9 & 23.6 / 7.2  &28.0 / 10.4  & 36.6 / 13.4\\\hline
GSP~\cite{pathak2018zero} &  31.7 / 13.4 & 25.1 / 9.6  &27.6 / 10.5  & 53.2 / 16.5\\\hline
NeoNav~\cite{wuneonav}&  33.4 / 12.5 &  19.7 / 6.1&  26.9 / 9.3 &  40.8 / 19.7 \\\hline
Ours&  42.6 / 23.6 &  \textbf{36.7 / 19.6}&  \textbf{40.6 / 21.8}&  \textbf{62.7 / 38.1} \\\hline
Ours-NoVal&  40.1 / 21.9 &  17.8 / 7.3 &  34.4 / 17.8 &  47.3 / 25.4 \\\hline
Ours-FroView&  34.8 / 11.2 &  17.6 / 5.0 &  28.0 / 9.2 &  48.8 / 16.0 \\\hline
Ours-NoGen&  38.8 / 22.0 &  28.8 / 15.4&  38.8 / 22.7&  58.4 / 35.2 \\\hline
Ours-VallinaGen &  \textbf{47.2 / 24.0} &  15.6 / 6.1 &  34.8 / 13.5 &  52.4 / 27.3 \\\hline
\end{tabular}}}\vspace{-6pt}
\end{table}

\begin{table*}
\centering
\caption{Navigation performance (SR and SPL in \%) on different input modalities from AI2-THOR with stop action.}
\label{tab:table1}\vspace{-6pt}
\scalebox{0.95}{
{\begin{tabular}{c|c|c|c|c|c|c}
\cline{1-7}
\hline
& Category & Kitchen & Living room & Bedroom  & Bathroom & Avg.\\
& & P=15.2\%& P=15.6\%& P=20.0\%&P=20.0\%&P=17.7\%\\\hline
\hline
&Random&  0.0 / 0.0 &  1.6 / 1.0&  2.0 / 1.1&  1.2 / 0.7 &1.2 / 0.7\\\cline{2-7}
Cross&Ours(RGB)&  42.6 / 23.6 &  \textbf{36.7 / 19.6}&  40.6 / 21.8&  \textbf{62.7} / 38.1 & 45.7 / 25.8\\\cline{2-7}
-scene&Ours(Semantic)&   \textbf{58.4 / 39.0} &  25.4 / 14.5&  \textbf{44.4 / 23.3}&  61.6 / \textbf{39.5}& \textbf{47.5 / 29.1}\\\hline
\hline
& Category & Kitchen & Living room & Bedroom  & Bathroom & Avg.\\
& & P=20.0\%& P=13.6\%& P=15.6\%&P=14.6\%&P=16.0\%\\\hline
\hline
&Random&  2.8 / 1.4 &  0.4 / 0.1&  1.6 / 0.1&  3.2 / 1.5 &2.0 / 1.0\\\cline{2-7}
Cross&Ours(RGB)&  46.6 / 26.1 &  \textbf{22.6} / 9.4&  39.0 / 21.1& 42.6 / 25.4& 37.7/20.5\\\cline{2-7}
-target&Ours(Semantic)&  \textbf{53.6 / 34.8} &  22.4 / \textbf{10.3}&  \textbf{43.2 / 23.1}&  \textbf{47.6 / 27.8} & \textbf{41.7 / 24.0}\\\hline
\end{tabular}}}\vspace{-6pt}
\end{table*}

\subsection{Additional Results}
\textbf{Scene category.} Table~\ref{tab:table2} presents the navigation performance on different scene categories, which is based on the $\{\emph{Unseen}~\emph{scenes},~\emph{Known}~\emph{targets}\}$ evaluation tasks from AI2-THOR in the main paper.
All methods consistently demonstrate impressive navigation performance in small scenes, e.g., kitchen and bathroom.
However, navigation in large rooms, e.g., living room, is much more challenging.

\begin{figure*}[thpb]
\begin{center}\vspace{-5pt}
\includegraphics[width=0.9\textwidth]{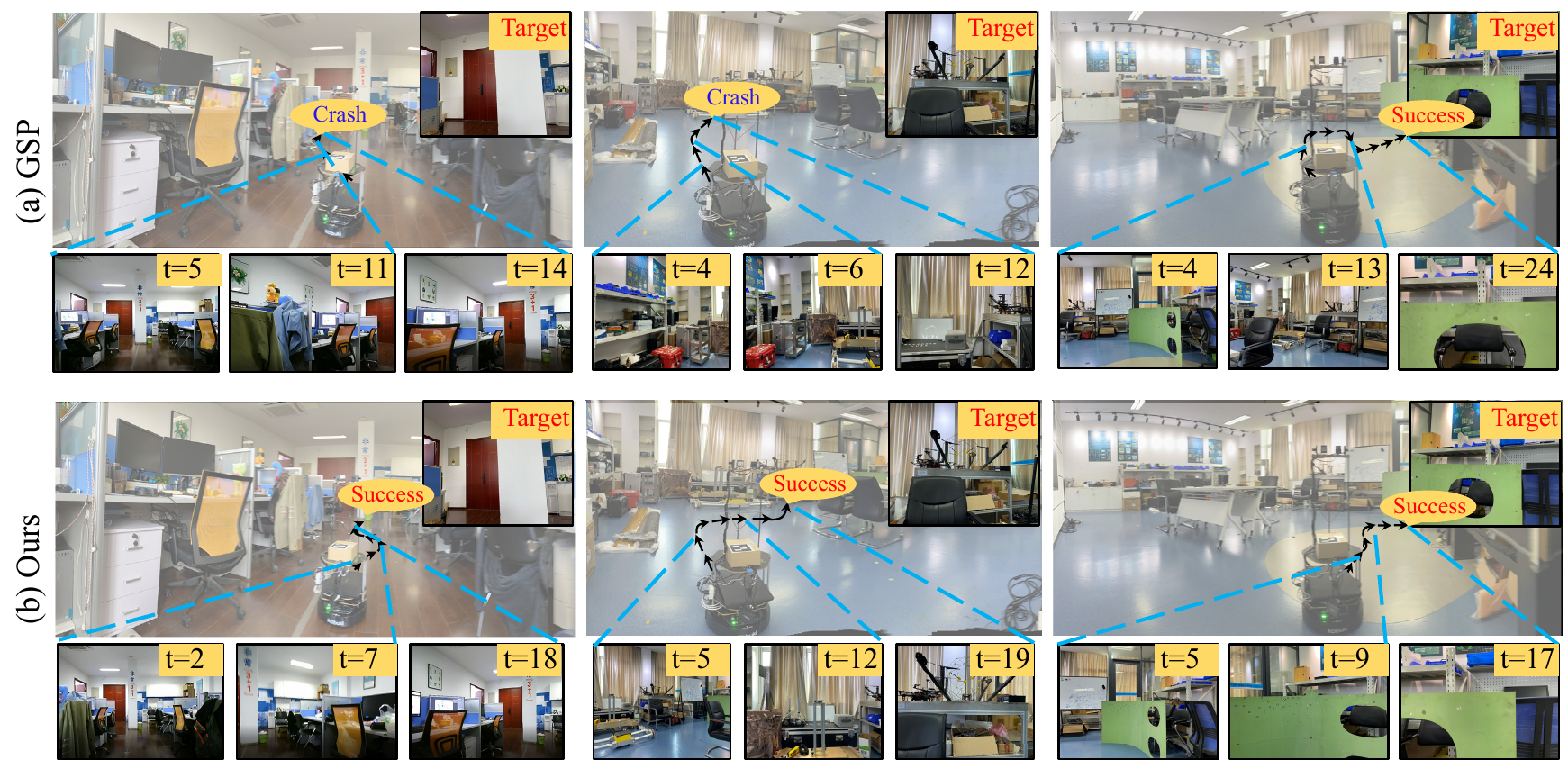}
\end{center}
   \caption{We compare our method with the GSP baseline. We illustrate the trajectories of the agent and the egocentric view of the agent at a few time steps. Our method finds the relatively safe ways towards the given navigation targets (e.g., avoiding collisions) and issues the stop action close to the targets.}\vspace{-12pt}
\label{fig:realnav}
\end{figure*}

\textbf{Input modality.} We conduct additional experiments of our method, where using semantic segmented images from AI2-THOR as inputs.
The training and testing setting is the same as the main paper.
In Table~\ref{tab:table1}, all navigation tasks are from the
evaluation of generalization on AI2-THOR in the main paper.
Although semantic segmented images are lossy compared to RGB, they do capture most of the important information for navigation, leading to substantial navigation performance improvement as expected.

\begin{table}[h]
\centering
\caption{Performance (SR) comparison of semantic navigation on the AVD test split.}
\label{tab:semg}\vspace{-6pt}
\resizebox{\columnwidth}{!}{
\begin{tabular}{l|c|c|c|c|c||c}
\cline{1-7}
\hline
Target label & Couch & Table & Refrigerator  & Microwave &TV & Avg. \\\cline{1-7}
\hline
\hline
TD-Semantic~\cite{mousavian2019visual} (Object)&  \textbf{80.0} &  38.0&   \textbf{68.0}&  38.0&  \textbf{44.0} &53.6\\\cline{1-7}
Ours (RGB)& 71.2 &  62.6&  51.0&  41.2&  39.6 & 53.1\\\cline{1-7}
Ours (Depth)&67.0&  \textbf{81.2}&  61.4&  \textbf{49.6} & 35.8 & \textbf{59.0}\\\cline{1-7}
\end{tabular}}\vspace{-6pt}
\end{table}

\textbf{Navigation on AVD.} We further present the best performances of our method and TD-Semantic~\cite{mousavian2019visual} over various target labels from AVD in Table~\ref{tab:semg}.
Both methods demonstrate better navigation performance when driven by relatively large targets (e.g., couch and refrigerator).
Our method (with depth input) outperforms TD-Semantic with semantic input by $5.4\%$
in terms of average success rate for all navigation tasks.
%Figure~\ref{fig:path} visualizes some typical success and failure cases of our method from AVD unseen scenes.

\textbf{Transfer to the real world.}
We compare our method with the GSP baseline based on navigation tasks from three real-world scenes. These scenes are significantly different from the household training scenes from AVD and the navigation targets have been never seen before the testing. Our method trained on the AVD, can transfer to these realistic scenarios and exhibit better navigation performance than GSP.
Figure~\ref{fig:realnav} qualitatively presents three navigation trajectories of GSP and our method, respectively. We also show the egocentric view of the agent at a few time steps during navigation.